\documentclass[runningheads]{llncs}
\usepackage{graphicx}
\usepackage{amsmath,amssymb}
\usepackage{array}
\usepackage[ruled,vlined]{algorithm2e}
\begin{document}
\title{Generative Data Augmentation for Vehicle Detection in Aerial Images}
\makeatletter{\renewcommand*{\@makefnmark}{}
\footnotetext{Workshop on Analysis of Aerial Motion Imagery (WAAMI 2020) in conjunction with 25th International Conference on Pattern Recognition (ICPR 2020)}\makeatother}
%
%
\author{Hilmi Kumdakcı \and
Cihan Öngün \and
Alptekin Temizel}
\authorrunning{H. Kumdakcı et al.}
%
\institute{Graduate School of Informatics\\Middle East Technical University, Ankara, Turkey
\email{\{hilmi.kumdakci,congun,atemizel\}@metu.edu.tr}}
\maketitle              
\begin{abstract}
Scarcity of training data is one of the prominent problems for deep networks which require large amounts data. Data augmentation is a widely used method to increase the number of training samples and their variations. In this paper, we focus on improving vehicle detection performance in aerial images and propose a generative augmentation method which does not need any extra supervision than the bounding box annotations of the vehicle objects in the training dataset. The proposed method increases the performance of vehicle detection by allowing detectors to be trained with higher number of instances, especially when there are limited number of training instances. The proposed method is generic in the sense that it can be integrated with different generators. The experiments show that the method increases the Average Precision by up to 25.2\% and 25.7\% when integrated with Pluralistic and DeepFill respectively.
\end{abstract}
\section{Introduction}
Computer vision applications on drone images are gaining importance with the need for automated analysis of increasing amounts of image data captured by drones. Object detection and image recognition tasks for vehicles and pedestrians, are at the heart of many applications such as surveillance. On the other hand, processing of aerial images comes with different challenges as the objects are relatively small and the data collection process is costly and done in uncontrolled environments, limiting the number of images in the datasets. Detection of small objects is generally handled by modifications on the detection network. For the lack of sufficient data, data augmentation techniques are used. Recently, Generative Adversarial Networks \cite{goodfellow2014generative} and Variational Autoencoders \cite{kingma2013auto} have been shown to generate realistic synthetic images. In this paper, we propose a data augmentation method using these generative networks. 

Typically, the number of annotated instances in aerial image datasets are relatively low compared to the common datasets such as COCO \cite{lin2014microsoft} and Pascal VOC \cite{Everingham10}. In this work, we aim to improve vehicle detection performance in aerial images by synthetic data augmentation on training images. The proposed framework consists of a generator, which generates candidate samples, and an independent detector which evaluates the quality of the samples. The framework is independent from the generator network models, and any generative network capable of image inpainting can be integrated.

\section{Related Work}
Data augmentation plays a key role in many data-driven neural network tasks such as object detection and object classification. Until recent advances, for visual tasks, data augmentation is mainly provided with spatial and intensity based modifications of images or instances. Even though it is not a guaranteed way of improving performance of a model, classical data augmentation methods are widely accepted as a primary solution to overcome scarcity of data because of ease of implementation. These techniques are mostly useful and applicable for almost all object detection tasks. These methods can be mainly grouped as geometrical and color based transformations. Each transformation method has its own advantages and might work better than other techniques under different conditions and for different data distributions.  When the main requirement for the model is robustness to varying illumination, color based illuminations might work perfectly even for small scale datasets. If the angle of the object changes during inference, geometric transformations might be much more helpful to improve the performance. Okafor et al. \cite{okafor_2017} used multi-orientation data augmentation to improve the classification of single aerial images of animals. They transform an input image to a new single image containing multiple randomly rotated versions the input image. Chen et al.\cite{chen_2018} used data augmentation for CNN-based people detection in aerial images. They applied image rotation, perspective transformation and border padding on aerial images. The main disadvantage of these methods is that they are not adding any distribution by discovering features of the data. If the task is to detect or classify classes when training set lacks necessary variety, this can not be considered as a good solution.

In addition to the traditional data augmentation methods, a number of recent approaches focus on data augmentation by copying existing object instances onto the existing training images \cite{lee2018context} \cite{Dvornik_2020} \cite{Dvornik_2018}. In \cite{lee2018context}, the ideal locations to place an object together with the best fitting pose of an instance for that scene are estimated. The idea is to provide locations for inserting objects into semantic maps using semantic segmented images to train generator network, since the purpose is placing generated objects visually plausible places in semantic maps, this idea is not defined as an augmentation for detection or classification tasks. In \cite{Dvornik_2020} \cite{Dvornik_2018}, a matching score between an  object and an image is calculated for augmentation. To prevent boundary artifacts, they use segmented annotated instances while choosing their data to be augmented. These set of methods are based on copying instances into different images without generative modelling.  PSGAN \cite{ouyang2018pedestriansynthesisgan} handles this contextual instance insertion problem by using a neural network based architecture having two discriminators and one generator. One of the discriminators is responsible for generating the instances and the other generates suitable patches for the generated instances. It uses spatial pyramid pooling layer to generate varying size instances. However, this approach has to deal with instances with artifacts due to the nature of generative networks.VS-GAN \cite{zheng2019using} also uses a 2 stage discriminator strategy for vehicle detection using least square loss in generator, and validates the augmentation with YOLOv3 and RetinaNet detectors. Since it is required to generate high quality synthetic samples, it was trained with a large scale dataset having car instances. DetectorGAN \cite{Liu_2019} has added a detector network into the generator-discriminator loop of PSGAN network to have more realistic outputs. This approach is highly dependent on training the discriminator branch which might require training parameters to be changed per subject. Another approach \cite{Han_2019} for medical studies keeps geometric and intensity information intrinsically while generating instances. In aerial images, \cite{milz2018aerial} to ours augments aerial images using image-to-image translation by conditional GANs. It is based on mapping the layout into another one while keeping instances which requires layout annotation. In our framework, we don't propose a new generator model but use generator and detector modules separately which is a generic approach for the problem. It prevents the generator overfitting and provides diverse and realistic augmentations. Also with the given parameters, cost and quality trade-off can be arranged.

\section{Background}
The proposed method consists of a generator network and a detector network. The generator network is expected to generate new instances that fits the given background and the detector network must be able to detect corresponding instances with bounding boxes. The networks must be generic networks that can work with different instance classes. For the experiments, we have selected Pluralistic Image Completion as generator network and Tiny YOLOv3 as detector network. In this section, we explain the selected networks and performance metrics for the experiments.

\subsection{Pluralistic}
Pluralistic Image Completion \cite{Zheng_2019} is an algorithm for one-to-many image completion tasks. In image completion, there is usually only one ground truth training instance per label which results in generated samples having limited diversity. To overcome this, Pluralistic uses two parallel paths, one is reconstructive and the other is generative, both are supported by GANs. The input images are partially masked to create synthetic holes. The algorithm generates diverse, realistic and reasonable images with completed holes. Let us define the original image as $I_g$, the partially masked image as $I_m$, and the complement image as $I_c$.  While the classical image completion methods attempt to reconstruct the ground truth image $I_g$ in a deterministic fashion from $I_m$, Pluralistic aims to sample from $p(I_c|I_m)$. The reconstructive path combines information from $I_m$ and $I_c$, which is used only for training. The generative path infers the conditional distribution of masked regions for sampling. Both of the paths follow Encoder-Decoder-Discriminator architecture. We used Pluralistic network as the generator model to generate car instances on given backgrounds.

\subsection{Tiny YOLOv3}
YOLO \cite{Redmon_2016} is a state-of-the-art, real-time object detection system. It uses a single optimized end-to-end network to predict bounding boxes and class probabilities directly from full images in a single pass. YOLOv1 introduced the concept of directly regressing object coordinates from the image instead of using region proposal networks such as Faster-RCNN \cite{Ren_2017}. Starting from YOLOv2, the system used anchor boxes for regressing object coordinates. YOLOv3 is trained with a different class prediction loss formula and makes detection at three different scales.

We used Tiny YOLOv3 \cite{yolov3} for experiments which is a smaller version for constrained environments. Since the aim of the proposed method is not to get the best detection performance among different models but to improve the performance of a base model by augmentation, we selected it considering its relatively low training/inference time. 

\subsection{Metrics} There are 2 common metrics for evaluating the detection performance. Intersection over union (IoU) is used for evaluating localization performance. It is the ratio of overlap (intersection) of predicted and ground-truth locations over union of predicted and correct locations (Eq. \ref{IOUEq}) where $B_g$ and $B_p$ are the ground truth and predicted bounding boxes of the object.
\begin{equation}
\label{IOUEq}
IoU(B_g,B_p) = \frac{|B_g \cap B_p|}{|B_g \cup B_p|}
\end{equation}
The result is between 0 and 1 indicating the ratio of correct prediction. If the prediction score is above the threshold, it is counted as a correct prediction.

Average Precision (AP) is the area under the precision-recall curve. It is commonly used to evaluate detection performance. Considering the correct predictions as true positives ($TP$), incorrect predictions as false positives ($FP$), and no predictions for an instance as false negatives ($FN$), we can formulate precision, recall and AP as in Eq. \ref{predrecallEq}:

\begin{align}
\label{predrecallEq}
precision = \frac{TP}{TP+FP} \qquad recall = \frac{TP}{TP+FN} \qquad AP = \int_{0}^1 p(r)dr
\end{align}

\section{Proposed Method}
\begin{figure}[t]
\centering
\includegraphics[width=0.4\textwidth]{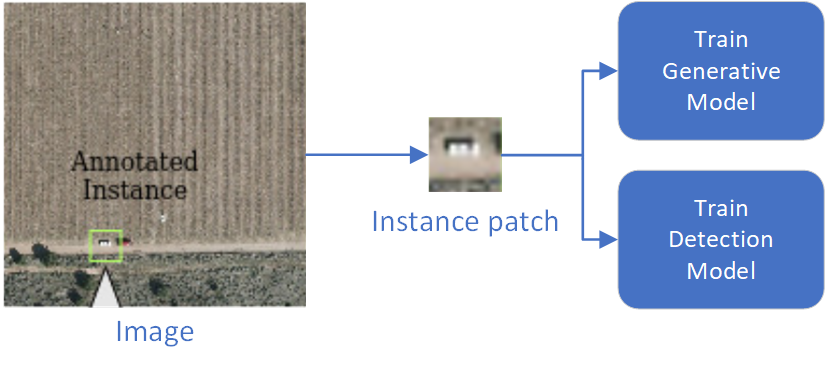} \\
\includegraphics[width=\textwidth]{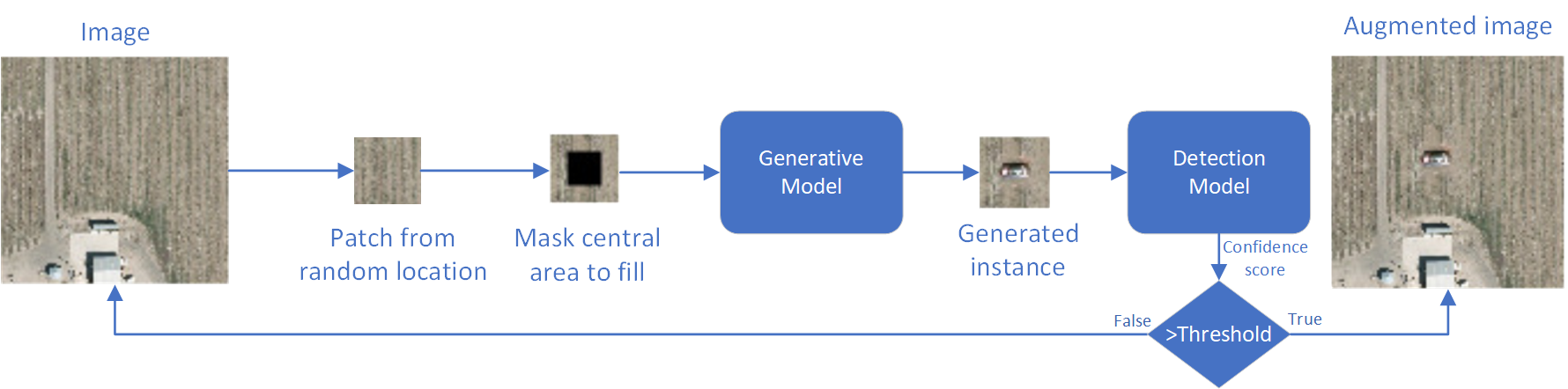}
\caption{Schematic of the proposed method. Top: Training stage, Bottom: Augmentation stage}
\label{fig:framework}
\end{figure}

The proposed method consists of 2 stages: training and augmentation.
Training stage involves independent training of a generative network and a detector network. At the augmentation stage, the generative network is used to generate new samples and the detector is used to assess the feasibility of these samples for augmentation. An augmented training set is formed using the samples which are deemed feasible after this assessment. This training set then can be used during training of a detector to improve the detection performance. The schematic of the proposed framework is shown in Fig. \ref{fig:framework}.

\subsection{Training Stage}
\label{sec:training}
The generator and detector networks are trained with image patches since the aim is to generate patches containing new instances. Both networks are fed with the same patches, extracted around the object instances from the training images. At the end of the training stage, the generator network learns to generate realistic instances on the given patches. The detector is trained with bounding box annotations and the best network parameters which has the highest average precision is selected for the augmentation stage.

\subsection{Augmentation Stage}
At this stage, the aim is to generate new object instances on the original images. First, 96x96 patches from random locations are extracted and their central 48x48 areas are masked. If the masked holes intersect with the existing instances, the patch is discarded and a new image is used for the patch extraction. The patches are fed into the trained generative model to generate synthetic object instances, which are expected to be located at the center of the patches. Then, the generated data is fed to the detector to evaluate whether the generated sample is acceptable to use for augmentation. For this purpose, the confidence score of the detector model, which reflects the confidence in identifying the generated instance, is used. The generated sample is accepted if the confidence score is higher than the predetermined threshold.  If the generated sample is not realistic or it has artifacts, it is expected to have a low confidence score. In this case, the augmentation stage starts over with the next image from the training set since the current image may not be suitable for augmentation. If the generated sample is accepted by the detector, the original image is augmented with the generated instance at the extracted patch location. The augmented set is formed by adding a predetermined number of new instances into the raw train set.

\section{Experimental Results}

\subsection{Setup}

\subsubsection{Dataset} We used Vehicle Detection in Aerial Imagery (VEDAI) \cite{razakarivony2016vehicle} dataset which has 1272 RGB color images at 1024x1024 resolution. All images are annotated with bounding boxes for labels such as car, boat and motorcycle. We have selected ``car" instances for our experiments. There are a total of 1377 car instances in the dataset. Object instances larger than 48x48 pixels (less than 3\% of all instances) are discarded as mentioned below. We divided the dataset as 500 and 772 for training and testing respectively. Only a part of the training set was used for experiments since we aim to improve the performance on small training datasets. 96x96 patches have been extracted around car instances from the training images which have a total of 490 car instances. The images have been down-scaled to the default input resolutions for performance evaluation.

\subsubsection{Patch Size Selection}
We analyzed the instance sizes in the dataset to select the appropriate patch size. The histogram of all instances in the dataset are shown on the left side of Fig. \ref{fig:hist}. As can be seen from this figure, 48x48 area covers more than 97\% real instances with the best quality generated samples and we selected the instance size as 48x48 considering the best coverage and quality. Larger areas would cover all instances, but, in that case, most patches would have unproportionately large background area compared to the area of generated instances. In \cite{Zheng_2019}, it is reported that image completion works the best when the original image is double the size of the generated part. Hence, we selected the patch size as 96x96.  The experiments show that generative models fail when the patch size is larger and generation time increases exponentially. Also, the boundary artifacts are more pronounced when patch sizes are smaller.

\subsubsection{Generator}
Pluralistic algorithm has been used with its original implementation\footnote{https://github.com/lyndonzheng/Pluralistic-Inpainting}. It uses Residual Blocks as the building block of the system. Each Residual Block consists of two convolutional layers and a residual connection with a convolutional layer. Encoder has 5 Residual Blocks. Decoder and Discriminator have 5 and 6 Residual Blocks respectively with an attention layer in the middle. Training of this network takes around 3 hours for 200 epochs on an NVIDIA GTX 1080 TI GPU.

\begin{figure}[t]
\centering
\includegraphics[width=0.4\textwidth]{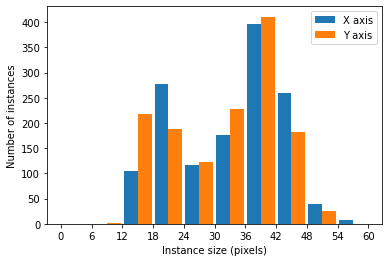}
\includegraphics[width=0.4\textwidth]{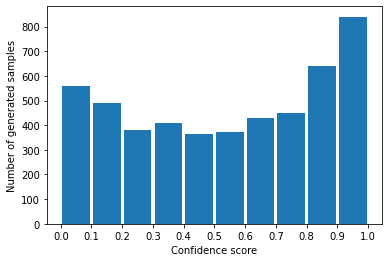} 
\caption{Left: Histogram of instance sizes for all instances in the training dataset. Right: Histogram of confidence scores for 5000 generated samples.}
\label{fig:hist}
\end{figure}

\begin{figure}[t]
\centering
0.0\hspace{0.8cm}0.1\hspace{0.8cm}0.2\hspace{0.8cm}0.3\hspace{0.8cm}0.4\hspace{0.8cm}0.5\hspace{0.8cm}0.6\hspace{0.8cm}0.7\hspace{0.8cm}0.8\hspace{0.8cm}0.9
\\
\includegraphics[width=0.09\textwidth]{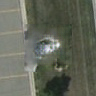}
\includegraphics[width=0.09\textwidth]{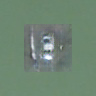}
\includegraphics[width=0.09\textwidth]{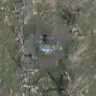}
\includegraphics[width=0.09\textwidth]{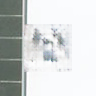}
\includegraphics[width=0.09\textwidth]{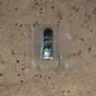}
\includegraphics[width=0.09\textwidth]{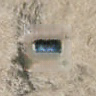}
\includegraphics[width=0.09\textwidth]{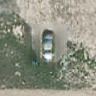}
\includegraphics[width=0.09\textwidth]{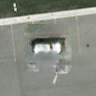}
\includegraphics[width=0.09\textwidth]{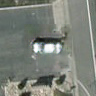}
\includegraphics[width=0.09\textwidth]{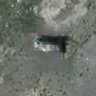}
\\
\includegraphics[width=0.09\textwidth]{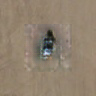}
\includegraphics[width=0.09\textwidth]{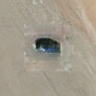}
\includegraphics[width=0.09\textwidth]{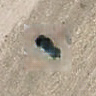}
\includegraphics[width=0.09\textwidth]{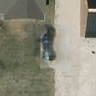}
\includegraphics[width=0.09\textwidth]{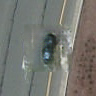}
\includegraphics[width=0.09\textwidth]{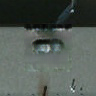}
\includegraphics[width=0.09\textwidth]{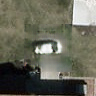}
\includegraphics[width=0.09\textwidth]{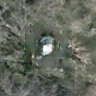}
\includegraphics[width=0.09\textwidth]{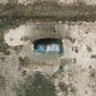}
\includegraphics[width=0.09\textwidth]{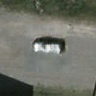}
\\
\includegraphics[width=0.09\textwidth]{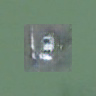}
\includegraphics[width=0.09\textwidth]{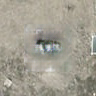}
\includegraphics[width=0.09\textwidth]{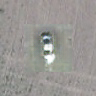}
\includegraphics[width=0.09\textwidth]{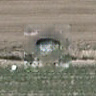}
\includegraphics[width=0.09\textwidth]{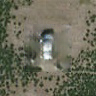}
\includegraphics[width=0.09\textwidth]{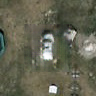}
\includegraphics[width=0.09\textwidth]{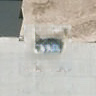}
\includegraphics[width=0.09\textwidth]{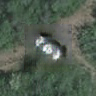}
\includegraphics[width=0.09\textwidth]{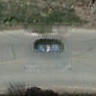}
\includegraphics[width=0.09\textwidth]{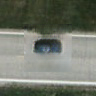}

\caption{Examples from generated samples with corresponding confidence scores.}
\label{fig:samples}
\end{figure}

The model is able to generate realistic and diverse outputs without mode collapse. To see the distribution of the generated samples, we generated 5000 samples and evaluated it with the detector. Some examples from generated samples are shown in Fig. \ref{fig:samples} and the histogram of confidence scores can be seen on the right side of Fig. \ref{fig:hist}. The equal-ranged histogram bins exhibit no large difference, indicating a good diversity over generated samples. Also it shows that the generator does not overfit to detector vulnerabilities.

\subsubsection{Detector}
Tiny YOLOv3 has been adopted as the detector module. It has 13 convolutional layers and the first six layers are followed by max-pooling layers. The official implementation has been used with the pretrained weights obtained by training on ImageNet \cite{imagenet_cvpr09} and default parameters for 96x96 resolution. The pretrained weights provides faster convergence, more stable training and better generalization. It also prevents overfitting due to single class training. We trained it for 2000 epochs and for 96x96 patches, the training takes less than 1 hour on the same GPU. For performance evaluation, it is trained with the default input size of the network (416x416) which takes around 7 hours. Workflow of the proposed method is summarized in Algorithm \ref{algo}.

\begin{algorithm}[t]
\SetKwInOut{Input}{input}\SetKwInOut{Output}{output}
\Input{Generative Network $g$, Detection Network $d$}
\SetKwProg{Train}{Training Stage}{:}{}
\SetKwProg{Aug}{Augmentation Stage}{:}{}
\SetAlgoNoLine
$w\leftarrow 96$ \tcp*[h]{Determined patch size}\\
\Train{}{
\SetAlgoLined
\For{$j\leftarrow 1$ \KwTo Number of instances}{ 
$p_j \leftarrow$ Extract $w \times w$ instance patches
}
Train Generative Network $g$ with extracted instance patches $p_{1...n}$\\
Train Detector Network $d$ with extracted instance patches $p_{1...n}$
}
\BlankLine
\SetAlgoNoLine
\Aug{}{
\SetAlgoLined
\For{$j\leftarrow 1$ \KwTo Augmentation count}{
Select an image to augment\\
$p \leftarrow $Extract patch from random location \\
Skip to another image if $p$ intersects with another instance\\
$p \leftarrow$ Mask the central $w/2 \times w/2$ area\\
$p' \leftarrow$ Generate instance with Generative Network $g(p)$\\
$o \leftarrow$ Evaluate generated instance with Detector Network $d(p')$ \\
\If{o $>$ Acceptance threshold}{
Augment the image with the generated instance
}
}
}
\caption{Workflow of proposed method}
\label{algo}
\end{algorithm}

\subsection{Results}
 
We first conducted an experiment to determine the best parameters for the networks by augmenting 500 images with 1000 generated instances. The result of this experiment can be seen in Table \ref{table:thresholdTest} for different confidence thresholds and IoU values. Confidence threshold decides if the generated sample is good enough to augment. When there is no threshold (= 0.0), the result is the worst as expected since all generated samples are augmented regardless of their quality. In this case, the process is also completed in 1000 iterations as there are no rejected samples. Considering the average of IoU values, the best threshold value is 0.9. At this threshold, it takes 5902 iterations to generate 1000 accepted instances, i.e. 1 sample is accepted from every $\sim$5.9 generated samples. It takes significantly long time and the diversity of the accepted samples is low. The augmented dataset does not contribute to the robustness and the generalization performance of the detector and may cause overfitting because of the lack of diversity. The second best threshold 0.4 provides a good balance between average precision and the number of iterations. It also has the best precision for the default IoU value (0.5). Considering the execution time and the diversity of the accepted samples, we selected 0.4 as the acceptance threshold.

\begin{table}[t]
\caption{Average Precision results for different confidence thresholds and IoU values where 500 images are augmented with 1000 instances. Augmentation iterations indicate the number of trials to reach 1000 accepted instances. }
\centering
\begin{tabular}{ >{\centering}p{2cm} | >{\centering}p{1.5cm} >{\centering}p{1.5cm} >{\centering}p{1.5cm} | >{\centering}p{1.5cm} | >{\centering\arraybackslash}p{2.5cm}}
Acceptance & \multicolumn{3}{c}{IoU} & & Augmentation \\
threshold & 0.2 & 0.5 & 0.7 & Average & iterations \\
\hline
\hline
0.0 & 56.20 & 36.11 & 7.26 & 33.19 & 1000 \\
0.1 & 55.55 & 40.67 & 7.97 & 34.73 & 1172 \\
0.2 & 55.42 & 39.51 & 9.23 & 34.72 & 1288 \\
0.3 & 56.23 & 39.65 & 8.76 & 34.88 & 1423 \\
0.4 & 56.38 & 44.16 & 10.31 & \textbf{36.95} & 1709 \\
0.5 & 56.04 & 44.05 & 10.08 & 36.72 & 1818 \\
0.6 & 56.16 & 42.29 & 10.81 & 36.42 & 2189 \\
0.7 & 57.63 & 43.41 & 9.57 & 36.87 & 2667 \\
0.8 & 57.64 & 43.15 & 9.90 & 36.90 & 3498 \\
0.9 & 57.08 & 43.81 & 10.57 & \textbf{37.15} & 5902 \\
\hline			
\end{tabular}
\label{table:thresholdTest}
\end{table}

There are two main factors for the evaluation of augmentation performance: number of training images and number of synthetic car instances. We have tested the system with varying number of training images to observe the performance of the system. The number of training images has been selected to be 200, 300, and 400, where they contain 251, 334, and 424 car instances respectively. The number of training instances are increased by adding 1 or 2 new instances per image by augmentation, i.e. when there are 400 training images, we added 400 or 800 synthetic object instances (cars) to approximately increase the existing number of instances by a factor of 2$\times$ and 3$\times$. Note that all of the synthetic object instances are placed on the original images, keeping the number of images the same. As per above example, there are still 400 images but they have higher number of object instances than they originally have.

\begin{figure}
\centering
\includegraphics[width=0.39\textwidth]{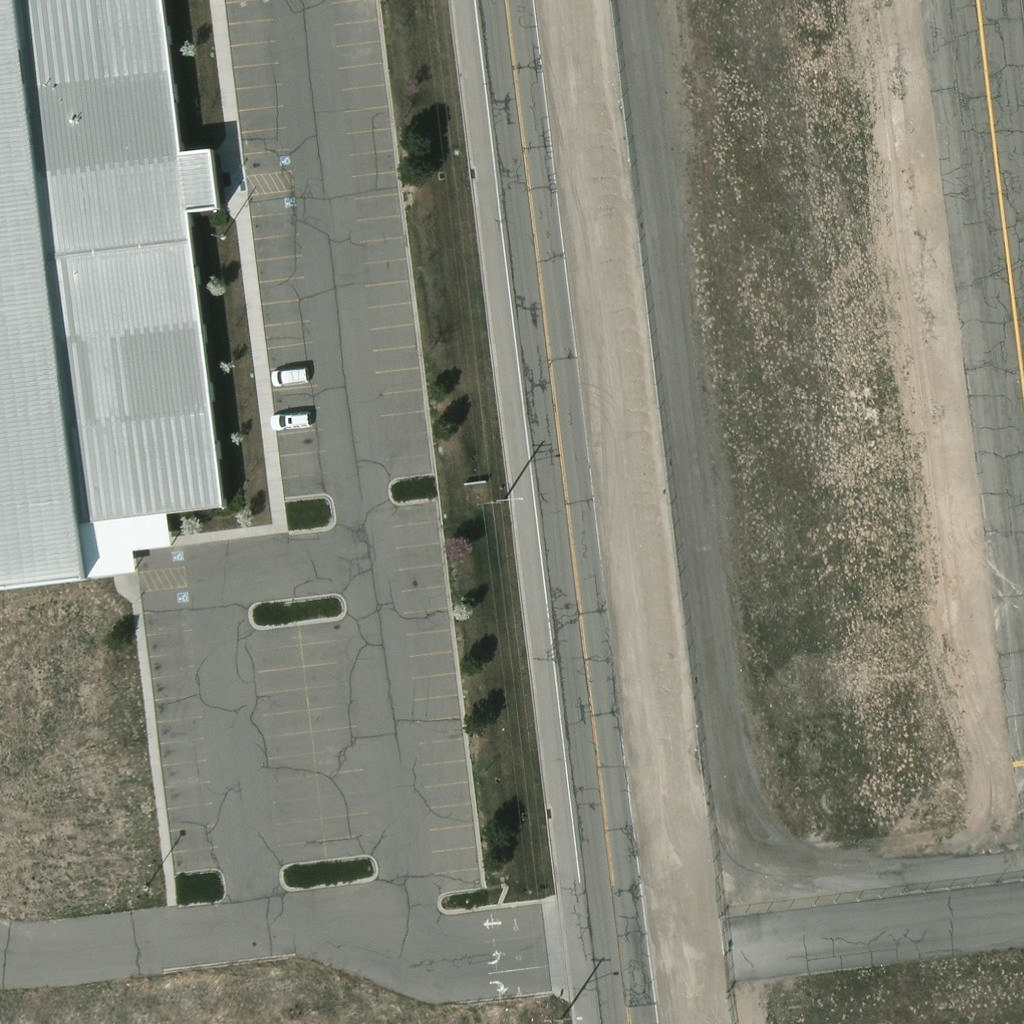}
\includegraphics[width=0.39\textwidth]{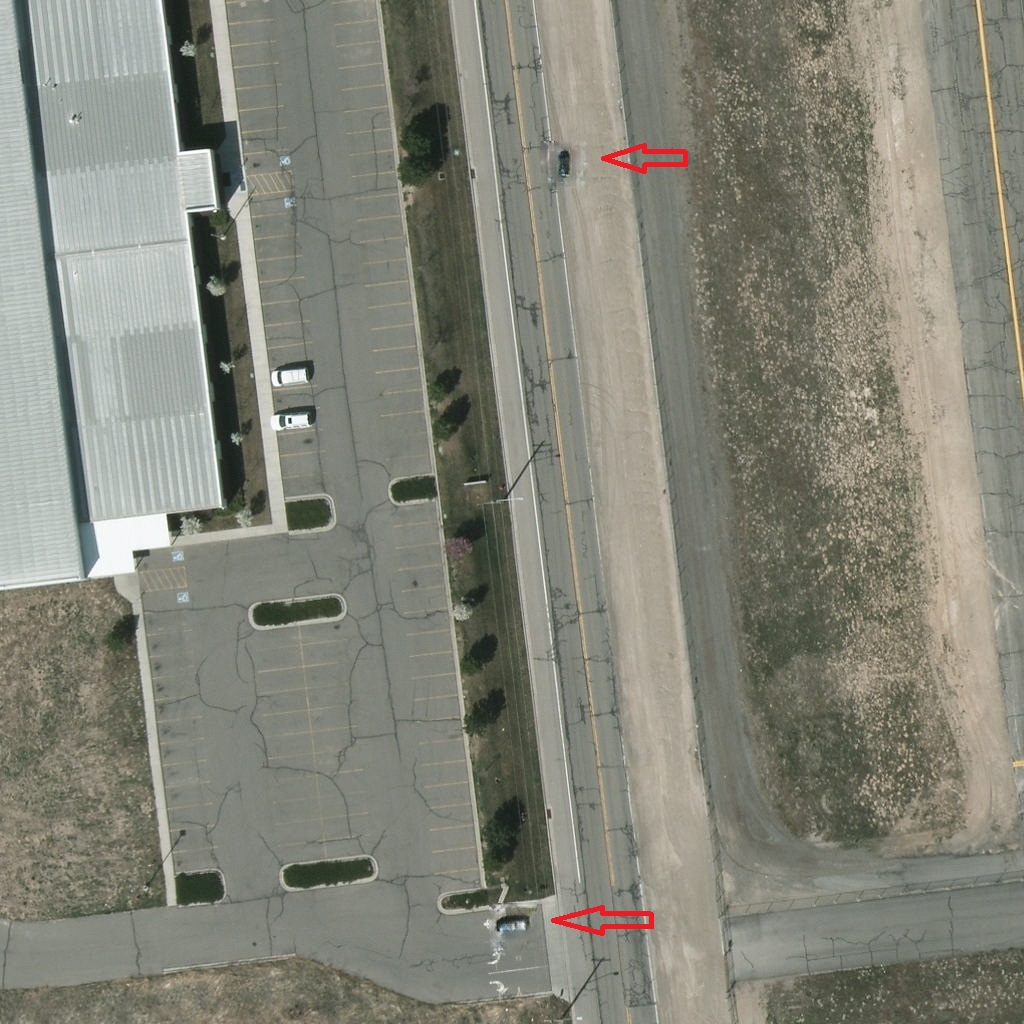}
\includegraphics[width=0.1935\textwidth]{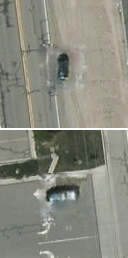}
\\
\includegraphics[width=0.39\textwidth]{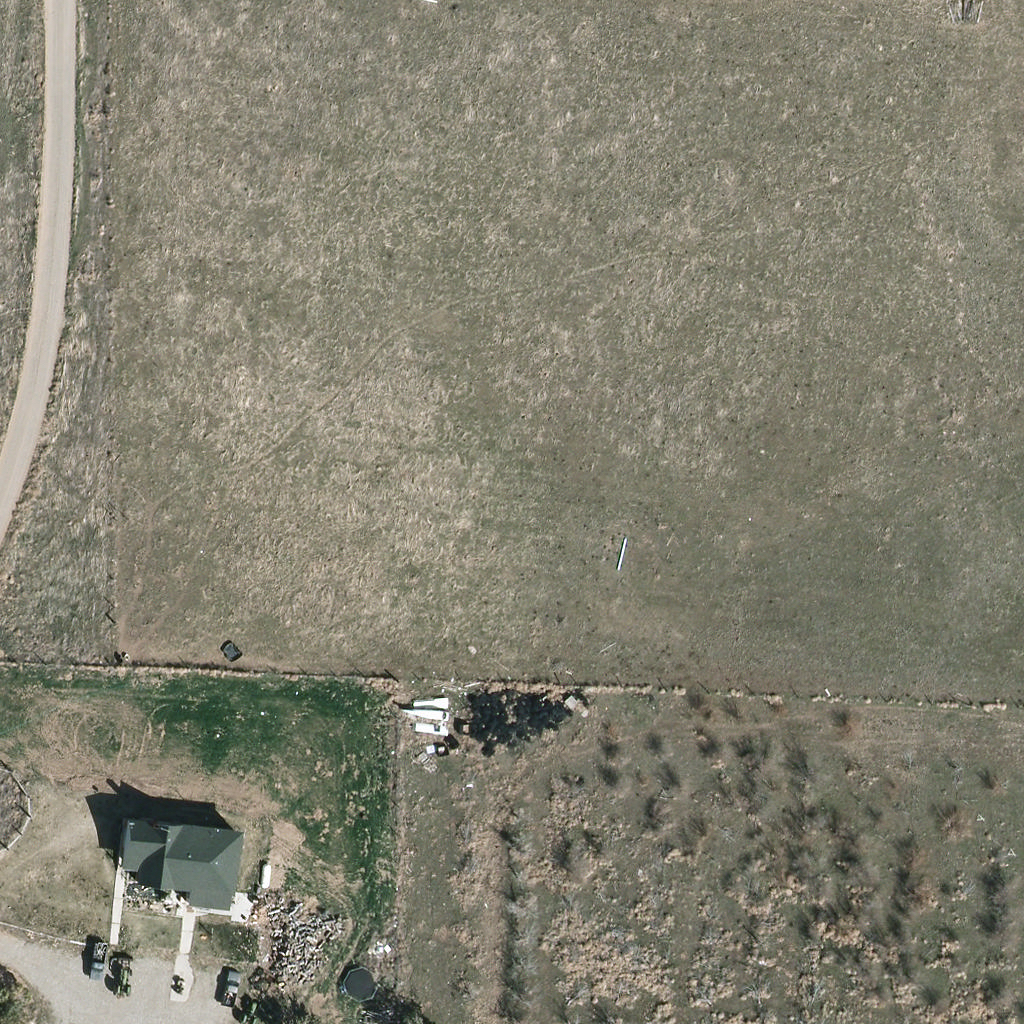}
\includegraphics[width=0.39\textwidth]{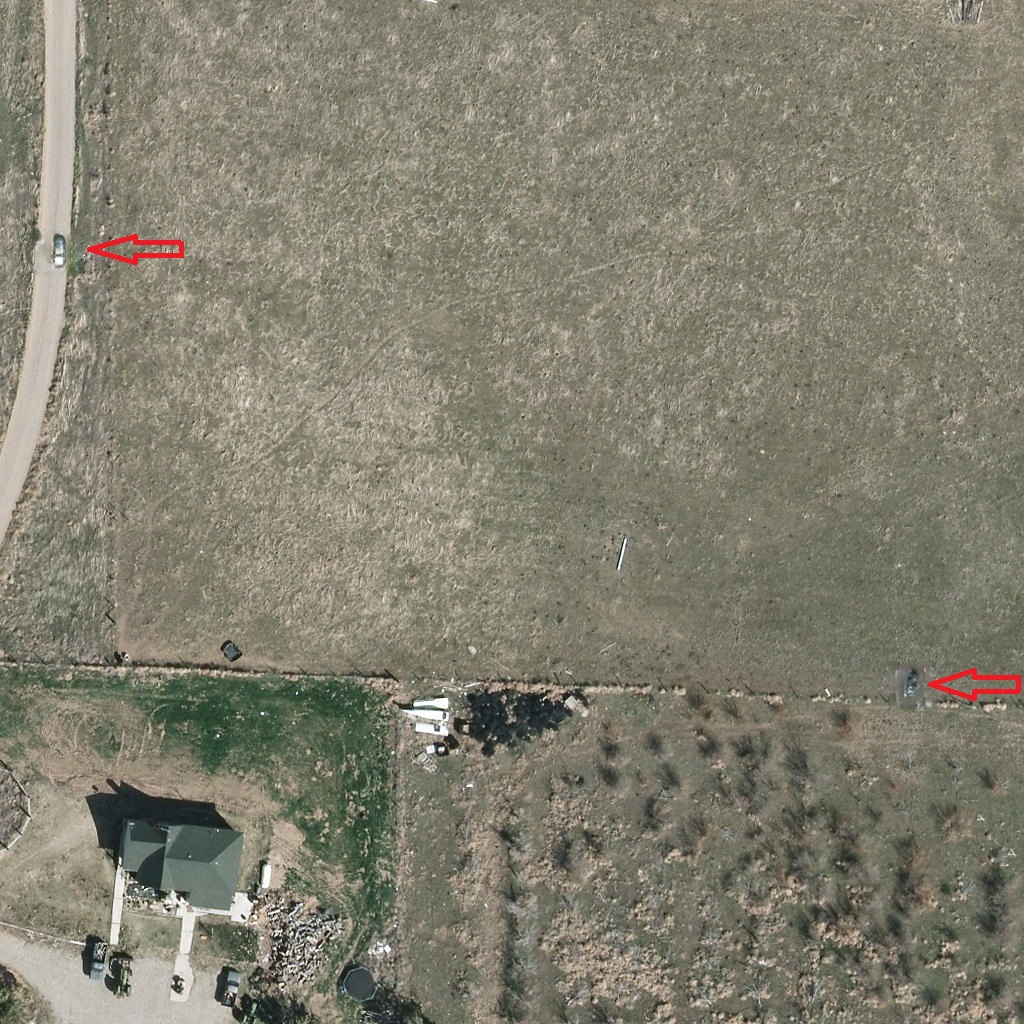}
\includegraphics[width=0.1935\textwidth]{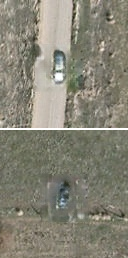}
\\
\includegraphics[width=0.39\textwidth]{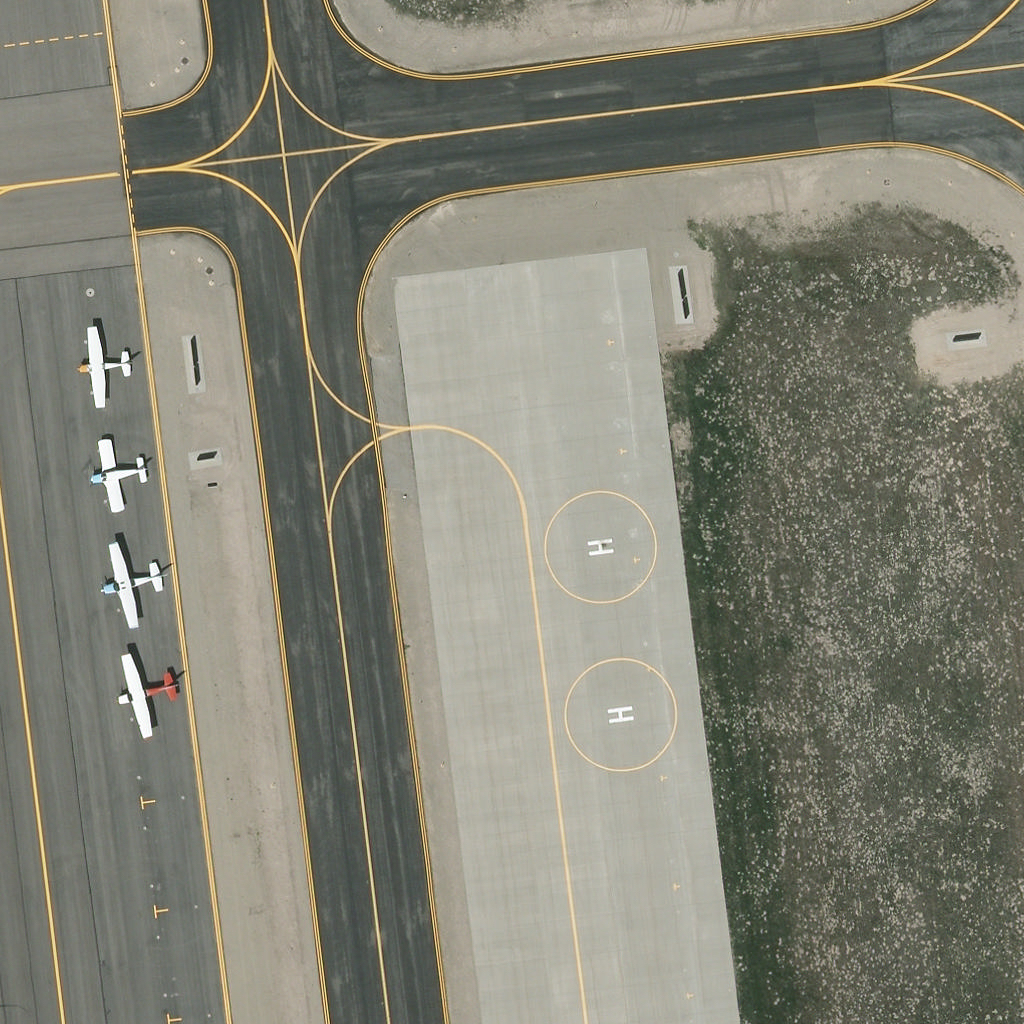}
\includegraphics[width=0.39\textwidth]{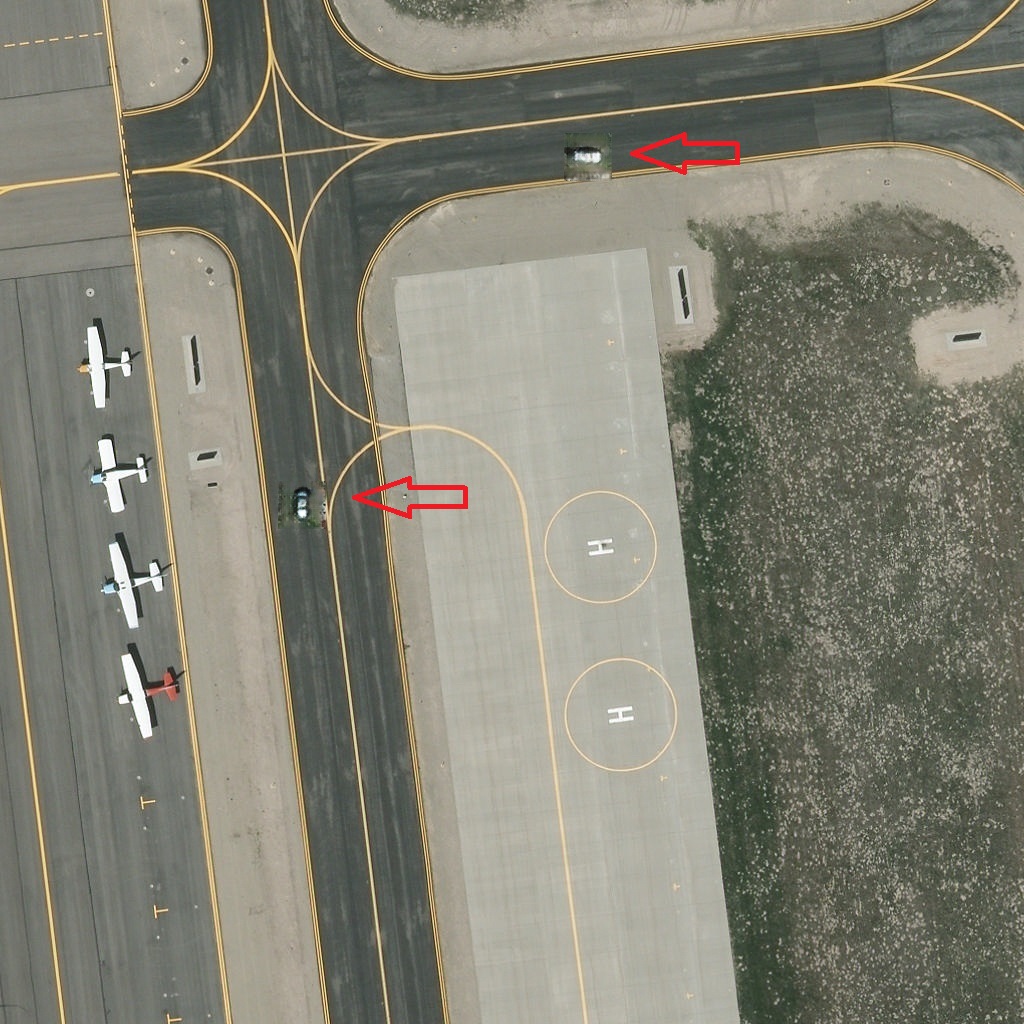}
\includegraphics[width=0.1935\textwidth]{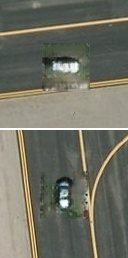}

\caption{Raw training images (left) are augmented with 2 separate car instances (middle). Augmented samples are highlighted with red arrows and their zoomed versions are shown on the right.}
\label{fig:augmented}
\end{figure}

The results shown in Table \ref{table:detectionTest} reveals that the performance of the detector network improves with the generative augmentation in all cases compared to the baseline (i.e. when there is no generative augmentation). Average Precision at IoU$ > 0.5$ performance increases by 17.8\%, 25.2\% and 16.7\% respectively when there are 200, 300 and 400 images in the dataset and they are augmented with 2 new instances per image. Examples from the augmented dataset can be found in Fig. \ref{fig:augmented} where it can be seen that the augmented instances are realistic, diverse and coherent with the rest of image. Both visual and quantitative results demonstrate that the proposed method can be used as an augmentation strategy to improve the performance especially when there are limited number of training images.

\begin{table}
\caption{Detection performances (AP) with the selected parameters.}
\centering
\begin{tabular}{>{\centering}p{1.5cm} >{\centering}p{2cm} >{\centering}p{1.5cm} >{\centering\arraybackslash}p{1.5cm} }
Dataset & Augmented & \multicolumn{2}{c}{IoU} \\
Images & Instances & 0.2 & 0.5 \\
\hline	
\hline	
200 & - & 49.85 & 31.43 \\
200 & 200 & 52.15 & 37.65 \\
200 & 400 & 51.71 & 37.04 \\
\hline
300 & - & 53.76 & 33.25 \\
300 & 300 & 55.39 & 39.76 \\
300 & 600 & 56.01 & 41.62 \\
\hline
400 & - & 55.46 & 36.14 \\
400 & 400 & 56.03 & 40.24 \\
400 & 800 & 56.02 & 42.18 \\
\hline			
\end{tabular}
\label{table:detectionTest}
\end{table}

\subsection{Additional Experiments}

\begin{figure}[t]
\centering
\includegraphics[scale=0.5]{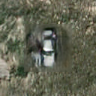}
\includegraphics[scale=0.5]{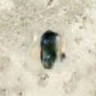}
\includegraphics[scale=0.5]{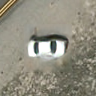}
\includegraphics[scale=0.5]{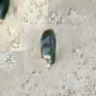}
\includegraphics[scale=0.5]{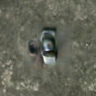}
\includegraphics[scale=0.5]{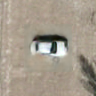}
\includegraphics[scale=0.5]{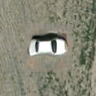}
\includegraphics[scale=0.5]{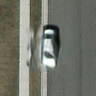}
\includegraphics[scale=0.5]{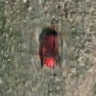}
\includegraphics[scale=0.5]{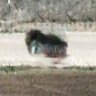}
\includegraphics[scale=0.5]{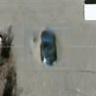}
\includegraphics[scale=0.5]{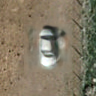}
\caption{Examples of generated samples with DeepFill.}
\label{fig:add_samples}
\end{figure}

We have conducted additional experiments to show that the proposed method can also be used with different generator models. DeepFill \cite{yu2018generative} is a generative inpainting network. The model is a feed-forward, fully convolutional neural network which can process images with multiple holes at arbitrary locations and with variable sizes. It has 2 autoencoders, the first one is for coarse inpainting and the following one is for refining the coarse generation. Two discriminators are used for local and global evaluation. It uses contextual attention and gated convolutions.  The official DeepFill implementation provided with the paper has been used for experiments. The proposed workflow has been integrated with DeepFill as generator instead of Pluralistic. Some examples of generated samples are shown in Fig. \ref{fig:add_samples} and the results can be seen in Table \ref{table:extraResults} for 0.9 detector acceptance threshold. Average Precision at IoU $> 0.5$ increases by 25.1\%, 25.7\% and 25.6\% respectively when there are 200, 300 and 400 images in the dataset and they are augmented with 2 new instances per image. It can be seen that the proposed method results in performance improvement for the detector model with similar gains to the variant with the Pluralistic, showing that the method can be used with different generative models. 
\begin{table}[b]
\caption{Detection performances (AP) when DeepFill is used as the generator model.}
\centering
\begin{tabular}{>{\centering}p{1.5cm} >{\centering}p{2cm} >{\centering\arraybackslash}p{1.5cm} }
Dataset & Augmented  & \\
Images & Instances & DeepFill \\
\hline	
\hline	
200 & - & 31.43 \\
200 & 200  & 35.96 \\
200 & 400  & 39.31 \\
\hline
300 & -  & 33.25 \\
300 & 300  & 38.93 \\
300 & 600  & 41.81 \\
\hline
400 & -  & 36.14 \\
400 & 400  & 38.73 \\
400 & 800  & 45.40 \\
\hline
\end{tabular}
\label{table:extraResults}
\end{table}

\section{Conclusion}

In this paper, we proposed a generative augmentation framework for improving the performance of object detection in aerial images for small datasets. The proposed method consists of a generator to generate instances at random locations and an independent detector to evaluate the feasibility of the generated instances for augmentation. The proposed method is generic and it can work with different generator and detector models. While we evaluated the method with Pluralistic and DeepFill, in the future, use of different generative models can be investigated. The experiments in this paper were based on a single instance class (i.e. cars). In the future, it can be extended to provide augmentation for multi-class detection problems by generating instances for multiple classes at once.

%
%
%
\bibliographystyle{splncs04}
\bibliography{egbib}

\end{document}